\documentclass[11pt]{article}

\usepackage[final]{acl}

\usepackage{times}
\usepackage{latexsym}
\usepackage{amsmath}
\usepackage{amssymb}
\usepackage{booktabs}
\usepackage{placeins}
\usepackage{xcolor}

\usepackage[T1]{fontenc}

\usepackage[utf8]{inputenc}

\usepackage{microtype}

\usepackage{inconsolata}

\usepackage{graphicx}
\usepackage{tcolorbox}
\usepackage{setspace}
\tcbuselibrary{breakable}
\usepackage{listings}

\title{MediEval: A Unified Medical Benchmark for Patient-Contextual and Knowledge-Grounded Reasoning in LLMs}

\author{Zhan Qu and Michael Färber\\
  TU Dresden and ScaDS.AI, Germany \\
  \texttt{\{zhan.qu, michael.faerber\}@tu-dresden.de} \\}

\begin{document}
\maketitle
\begin{abstract}

Large Language Models (LLMs) are increasingly applied to medicine, yet their adoption is limited by concerns over reliability and safety. Existing evaluations either test factual medical knowledge in isolation or assess patient-level reasoning without verifying correctness, leaving a critical gap. We introduce MediEval, a benchmark that links MIMIC-IV electronic health records (EHRs) to a unified knowledge base built from UMLS and other biomedical vocabularies. MediEval generates diverse factual and counterfactual medical statements within real patient contexts, enabling systematic evaluation across a 4-quadrant framework that jointly considers knowledge grounding and contextual consistency. Using this framework, we identify critical failure modes, including hallucinated support and truth inversion, that current proprietary, open-source, and domain-specific LLMs frequently exhibit. To address these risks, we propose Counterfactual Risk-Aware Fine-tuning (CoRFu), a DPO-based method with an asymmetric penalty targeting unsafe confusions. CoRFu improves by +16.4 macro-F1 points over the base model and eliminates truth inversion errors, demonstrating both higher accuracy and substantially greater safety.


\end{abstract}

\section{Introduction}
Large Language Models (LLMs) have demonstrated remarkable capabilities across diverse domains, with medicine being among the most high-impact areas of application. In clinical contexts, LLMs have been explored for tasks such as summarizing electronic health records (EHRs), generating discharge instructions, providing clinical decision support, and answering medical questions \citep{rajpurkar2022ai, singhal2025toward, moor2023foundation, singhal2023large, pal2022medmcqa}. Their appeal lies in the ability to integrate unstructured text with medical knowledge, potentially reducing documentation burden and assisting clinicians in decision-making.

Translating research prototypes into real-world deployment hinges critically on reliability and safety. Unlike generic NLP tasks, medical reasoning requires not only factual correctness but also contextual grounding in patient-specific data while adhering to verified medical knowledge. Errors in this setting are not mere degradations in performance but risks that can directly translate into patient harm \citep{thirunavukarasu2023large, yang2023large, haltaufderheide2024ethics}. 

A critical challenge is that LLMs often fail to apply medical knowledge consistently within the heterogeneous and noisy context of patient records \citep{zhou2025evaluating}. For example, a model may state that metformin, a therapy for type 2 diabetes, is contraindicated in severe renal impairment, yet fail to apply this knowledge when the condition appears in a noisy and heterogeneous patient record. The cause of such errors may be that current models are trained to recall facts in isolation rather than to integrate them with diverse patient information. Such inconsistencies expose a critical gap between \emph{knowing} medical facts and \emph{using} them safely.

\begin{figure*}[!ht]
    \centering
    \includegraphics[width=1\linewidth]{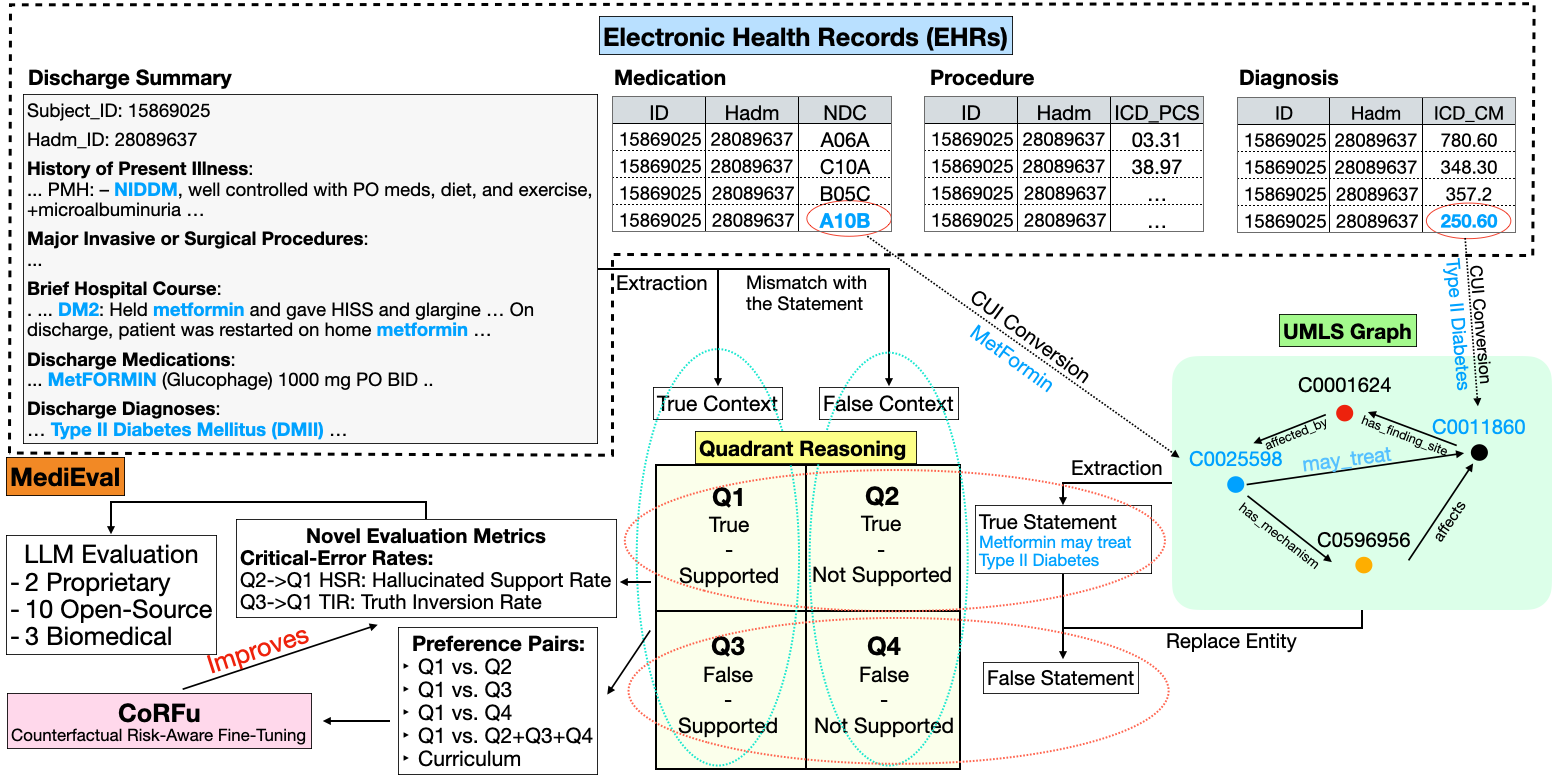}
    \caption{For each admission, MediEval uses both unstructured discharge summaries and structured EHR tables (diagnoses, procedures, and medications). First, clinically relevant sections are extracted from the discharge summary to form the patient context. In parallel, structured medical codes are semantically normalized into UMLS Concept Unique Identifiers (CUIs), which are then linked in an ontology graph to identify clinically meaningful relations, including multi-hop connections when needed. Based on these relations, MediEval constructs candidate statements and evaluates them under a four-quadrant framework defined by factual correctness with respect to biomedical knowledge (True vs. False) and contextual grounding with respect to the extracted patient context (Supported vs. Not Supported): Q1 = True–Supported, Q2 = True–Unsupported, Q3 = False–Supported, and Q4 = False–Unsupported. The resulting benchmark is used to evaluate LLMs and to support Counterfactual Risk-Aware Fine-tuning (CoRFu). Blue text marks the entities or statements extracted from the real patient example.}
    \label{fig:overview}
\end{figure*}

Existing evaluation paradigms only partially address this gap. Medical data are uniquely challenging because patient records are heterogeneous with free-text clinical notes and tabular entries coded in different systems for diagnoses, procedures, and medications. Medical knowledge is hierarchical and ontology-driven (e.g., UMLS, SNOMED CT, RxNorm), but its large scale, noise, and limited cross-vocabulary connectivity make consistent reasoning difficult. Benchmarks based on EHRs assess the ability to extract or reason over structured data, but often reduce the task to retrieval or serialization without verifying medical soundness \citep{lovon2025evaluating}. In contrast, knowledge-based evaluations test whether LLMs can handle logical transformations of medical facts \citep{zhou2025reliable, sung-etal-2021-language}, but do not connect reasoning to real patient contexts. The field thus lacks a unified framework that probes whether LLMs can (i) remain faithful to medical knowledge and (ii) apply it consistently to individual patient records.

In this paper, we address this gap with \textbf{MediEval} (Figure~\ref{fig:overview}), a benchmark and evaluation framework that links real patient records (MIMIC-IV) \citep{johnson2023mimic} with a unified biomedical knowledge base built from UMLS \citep{bodenreider2004unified}, SNOMED CT \citep{donnelly2006snomed}, and RxNorm \citep{liu2005rxnorm,nelson2011normalized}. To ensure rigorous construction, MediEval constructs evaluation statements by applying graph-guided substitutions and recombinations within biomedical ontologies, with plausibility checks to ensure that true cases remain clinically valid while false ones are realistic and challenging. These statements are embedded in real patient contexts, enabling evaluation of quadrant-level reasoning, defined as \emph{True/False} with respect to medical knowledge and \emph{Supported/Unsupported} with respect to the patient record. To capture safety-critical errors, we introduce two new metrics: Hallucinated Support Rate (HSR) and Truth Inversion Rate (TIR). We then conduct a comprehensive evaluation of proprietary, open-source (general-purpose), and biomedical LLMs under this unified protocol, revealing consistent weaknesses across quadrants.

To mitigate the identified risks, we introduce \textbf{Counterfactual Risk-Aware Fine-tuning (CoRFu)}, a DPO-based method that uses quadrant-structured preference pairs and an asymmetric penalty to push models to prefer “True+Supported” responses over safety-critical confusions (e.g., “False+Supported”). This risk-aware, counterfactual approach targets precisely the failure modes that conventional preference optimization overlooks. Our code is publicly available\footnote{\href{https://github.com/ZhanQu945/MediEval}{https://github.com/ZhanQu945/MediEval}}. Our key contributions are:




\begin{itemize}
\setlength{\itemsep}{0pt}
    \item \textbf{MediEval Benchmark.} A unified benchmark linking EHRs with biomedical ontologies to test both factual grounding and patient consistency, with novel safety evaluation metrics.  
    \item \textbf{Comprehensive Evaluation.} Cross-model evaluation of proprietary, open-source, and biomedical LLMs under identical settings, revealing systematic quadrant-level weaknesses and safety-critical error patterns.
    \item \textbf{Counterfactual Risk-Aware Fine-Tuning (CoRFu).} A risk-aware extension of DPO with asymmetric penalties and a quadrant-aware approach that reduces safety-critical errors in medical reasoning tasks.  
\end{itemize}

\FloatBarrier
\section{Related Work}

A growing body of work investigates whether LLMs can handle complex information in EHRs. One study focused on data serialization and retrieval, with performance sensitive to prompt design and feature selection \citep{lovon2025evaluating}. Another work has explored long-context modeling, showing some architectures can process entire patient timelines with $>$10k events and achieve greater robustness to irregular temporal patterns \citep{wornow2025context}. Additional approaches integrate structured and unstructured data, for instance, by using small auxiliary models as “knowledge triggers” to support tabular prediction \citep{yan2025small} or by designing code-aware representations that capture semantic and temporal structure in diagnosis prediction \citep{tan2025boxlm}. More recent work has explored longitudinal clinical reasoning through structure-aware retrieval and geometric representations of patient trajectories, enabling more clinically consistent next-event forecasting \citep{qu2026grail}. Related efforts model evolving patient care as a continuous temporal reasoning problem, using structured memory and agentic context management to support long-horizon inference over streaming EHRs \citep{qu2026trace}. These studies demonstrate progress in surfacing and encoding patient information, but whether model outputs are consistent with established clinical knowledge remains underexplored.


Another line of research investigates the reliability of the medical knowledge encoded in LLMs. Dynamic probing methods show that models often achieve low joint accuracy when facts are rephrased \citep{zhou2025reliable}, while multifaceted evaluations reveal brittle performance on tasks requiring comparison, verification, or rectification \citep{zhou2024multifaceteval}. Broader studies also report steep declines when moving from factual recall to scenario-based reasoning as cognitive complexity increases \citep{zhou2025evaluating}. However, even when scenarios are considered, these evaluations do not disentangle whether errors arise from factual misunderstanding, failures of contextual grounding, or both.


Several approaches have sought to improve the factual reliability of LLMs by grounding them in structured biomedical knowledge. These include triplet generation with knowledge graph verification \citep{su2025kgarevion}, alignment with biomedical knowledge graph embeddings \citep{sakhovskiy2025bali}, and graph-based retrieval augmentation for medical question-answering \citep{wu-etal-2025-medical}. Scholars have also critiqued the evaluation paradigms: some argue that multiple-choice questions reward shallow pattern recognition rather than genuine medical knowledge \citep{griot2025pattern}, while others stress that medical benchmarks must prioritize construct validity to remain clinically meaningful \citep{alaa2025position}. Automatic evaluation frameworks, such as AutoMedEval \citep{zhang-etal-2025-automedeval}, further highlight the need for scalable yet reliable assessment pipelines.

\section{Methodology}
\label{sec:methodology}

\subsection{MediEval Framework: Data Construction}

MediEval is a benchmark constructed from the MIMIC-IV database (Medical Information Mart for Intensive Care IV) \citep{johnson2023mimic} in combination with biomedical ontologies such as UMLS \citep{bodenreider2004unified}. It is designed to evaluate whether large language models (LLMs) can perform clinical inference by verifying whether medical statements are factually correct and properly grounded in patient records. MediEval integrates structured EHR tables, unstructured clinical notes, and ontology-derived knowledge into a unified dataset of context–statement pairs annotated with correctness and support labels (see Figure~\ref{fig:overview}).

\paragraph{Electronic Health Records (EHRs) as Foundation.}  
MIMIC-IV is a large-scale publicly available collection of de-identified electronic health records (EHR) comprising 546{,}028 hospital admissions for 223{,}452 unique individuals. Two modalities are utilized: (i) structured tabular data (lists of diagnoses, procedures, and medications recorded with ICD-CM, ICD-PCS, and NDC codes), and (ii) unstructured discharge summaries, which typically contain about 22 sections such as \emph{Brief Hospital Course}, \emph{History of Present Illness}, \emph{Major Surgical or Invasive Procedures}, and \emph{Discharge Diagnoses}. 


For each admission $a$, we define the structured event set and the discharge summary section set as
\begin{align}
S_a  &= \{\, \mathcal{Z}^{\text{diag}}_{i_a},\; \mathcal{Z}^{\text{proc}}_{j_a},\; \mathcal{Z}^{\text{med}}_{k_a} \,\}, \label{eq:Sa}\\
DS_a &= \{\, s^{(1)}_a,\, s^{(2)}_a,\, \ldots,\, s^{(m_a)}_a \,\}. \label{eq:DSa}
\end{align}

Here $\mathcal{Z}^{\text{diag}}_{i_a}$, $\mathcal{Z}^{\text{proc}}_{j_a}$, and $\mathcal{Z}^{\text{med}}_{k_a}$ denote the lists of diagnosis, procedure, and medication codes for admission $a$, containing $i_a$, $j_a$, and $k_a$ elements respectively, while $DS_a$ represents the set of $m_a$ segmented sections from the discharge summary. See Figure~\ref{fig:overview} for a real sample from MIMIC-IV.


\paragraph{Step 1: Context Extraction.}  
Discharge summaries are divided into multiple sections, not all of which are equally informative. 
For MediEval, we define the evaluation context $\mathcal{C}_a \subset DS_a$ as the subset of discharge summary sections that best capture patient trajectory and clinical decision-making. In practice, we select sections covering diagnoses, procedures, treatments, medical history, and hospital course, as these jointly provide the most comprehensive evidence. Importantly, the reasoning that connects these entities is not explicitly structured but embedded in free-text narratives, requiring models to infer relations such as diagnosis–medication from natural language.


\paragraph{Step 2: Semantic Normalization of Medical Codes.} Electronic health records employ heterogeneous coding systems: diagnoses and procedures in MIMIC-IV are stored using ICD-9/10, while medications are represented with National Drug Codes (NDC). To reason across modalities, these codes must be projected into a unified semantic space. We achieve this by leveraging the Unified Medical Language System (UMLS) \citep{bodenreider2004unified}, which integrates over 60 families of biomedical vocabularies and provides explicit mappings between them. Specifically, ICD codes are aligned with SNOMED CT concepts, NDC codes are aligned with RxNorm drug identifiers, and all are normalized into Concept Unique Identifiers (CUIs) that serve as universal nodes for cross-system integration. Formally,
\begin{equation}
f : z \mapsto \text{CUI}(z), \; U_a = \{ f(z) \mid z \in S_a \}.
\end{equation}

Here, $z$ denotes a raw code that is being mapped to a $\text{CUI}(z)$, and $U_a$ is the set of normalized CUIs for admission $a$. This normalization ensures that diagnoses, procedures, and medications from different coding systems can be compared, linked, and reasoned over in a unified space.

\paragraph{Step 3: Ontology Graph Construction and Relation Extraction.}  
From the normalized concepts across the cohort, we construct a semantic graph $G=(V,E)$, where $V$ are CUIs and $E$ are relations drawn from the UMLS Metathesaurus, including \emph{treated\_by}, \emph{has\_associated\_procedure}, and \emph{is\_a} etc. A dictionary mapping synonyms and abbreviations to CUIs supports robust entity matching in text. Many clinically valid associations are not encoded as direct links but instead mediated through intermediate nodes such as therapeutic classes, anatomical structures, or related conditions. To capture these, we allow multi-hop traversal in $G$, limited to paths of length at most three in order to preserve clinical plausibility.

Formally, for concepts $h,t \in V$, let $d(h,t)$ be their hop distance; we define their relations as
\begin{equation}
R(h,t) = \{\, r \mid (h,r,t) \in G,\; d(h,t) \leq 3 \,\}
\end{equation}

\textit{Direct relations.}  
The semantic graph captures clinically meaningful associations as single-edge links. For instance, it encodes that the antihypertensive drug \emph{Lisinopril} (CUI C0023861) is linked to the condition \emph{Hypertension} (CUI C0020538) through a \emph{treated\_by} relation. Similarly, it represents that the procedure \emph{Coronary Angioplasty} (CUI C0001979) is associated with the diagnosis \emph{Coronary Artery Disease} (CUI C0010054) via a \emph{has\_associated\_procedure} relation. 

\textit{Multi-hop inference.}  
Not all clinically valid relations are encoded as direct edges between specific entities in UMLS. For example, many drug–disease links are only represented at the therapeutic class level rather than for individual drugs. As a result, the relation between \emph{Lisinopril} and \emph{Hypertension} may absent as a direct edge. Instead, the graph encodes that Lisinopril (CUI C0023861) belongs to the class \emph{Angiotensin-Converting Enzyme Inhibitors} (CUI C0003028) via an \emph{is\_a} relation, and that this class as a whole is linked to \emph{Hypertensive disease} (CUI C0020538) via a \emph{treats} relation:
\begin{equation*}
\text{(C0023861)} \xrightarrow{\text{is\_a}} 
\text{(C0003028)} \xrightarrow{\text{treats}} 
\text{(C0020538)}.
\end{equation*}
By traversing this two-hop path, the system can infer the treatment relation that is missing at the leaf level. Such multi-hop reasoning is essential for capturing clinically valid associations that are indirectly encoded in UMLS. 



In summary, $U_a$ provides the admission-specific set of normalized concepts, while $G$ encodes both direct and inferred relations among these concepts. Together they form the foundation for structured fact extraction in MediEval.

\paragraph{Step 4: Quadrant-based Statement Construction.}  
For each fact $(h,r,t)$, we generate a set of samples covering the four MediEval quadrants:
\begin{equation}
\mathcal{Y}_a(h,r,t) = \{ y^{Q1}, y^{Q2}, y^{Q3}, y^{Q4} \}.
\end{equation}

Quadrants are defined by the cross of factual correctness (true vs. false) and contextual grounding (supported vs. unsupported), as summarized in Figure~\ref{fig:overview}. This design ensures that evaluation covers both straightforward comprehension and more subtle failure modes such as plausible contradictions, relational mismatch, and erroneous information.

\begin{figure}[!ht]
    \centering
    \includegraphics[width=1\linewidth]{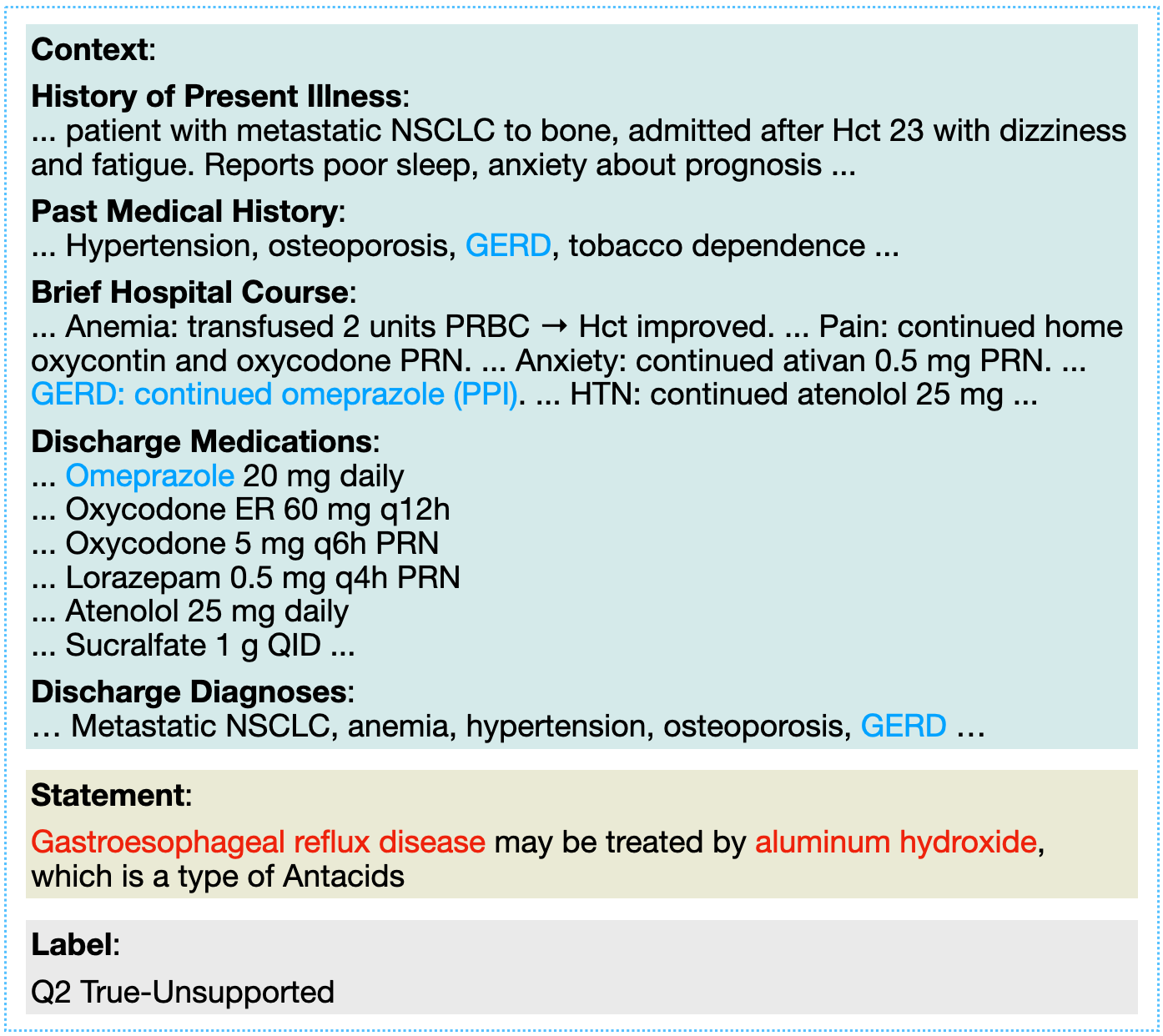}
    \caption{Example of statement verification against patient records (Quadrant 2: True-Unsupported). The statement is medically correct, yet in this case, GERD is treated with omeprazole, a different class of medication, so the statement is not supported by the context.}
    \label{fig:example}
\end{figure}

The quadrants are constructed as follows.  
\emph{Q1 (True–Supported, Supported Fact)} represents the baseline case: a correct fact directly attested in the patient context $\mathcal{C}_a$.  
\emph{Q2 (True–Unsupported, Plausible Contradiction)} is generated by replacing an entity with a closely related alternative from the ontology graph $G$ (e.g., another entity sharing the same parent concept) that is absent from $\mathcal{C}_a$, yielding a statement that is medically correct in general but unsupported for this patient (see Figure~\ref{fig:example} for an example). 
\emph{Q3 (False–Supported, Relational Mismatch)} recombines entities from distinct true facts, creating a false relation even though all entities appear in $\mathcal{C}_a$, thereby testing whether a model can resist misleading but superficially plausible associations.  
Finally, \emph{Q4 (False–Unsupported, Erroneous Information)} introduces a semantically distant distractor entity from $G$, producing a statement that is both false and unsupported.  


This procedure requires ontology-guided substitutions, recombinations, and plausibility checks to guarantee that true statements remain clinically valid while false ones are both plausible and challenging. The resulting dataset integrates curated narrative context, ontology-based code normalization, and graph reasoning into a unified framework for rigorous evaluation of factual reliability and context adherence. See Figure~\ref{fig:example} and Appendix~\ref{sec:appendixa} for real samples of each quadrant.

\subsection{MediEval Framework: LLM Evaluation}
\label{sec:evaluation}

\paragraph{Quadrant Formulation as a Four-Way NLI Task.}  
Given a patient context $c$ and a candidate medical statement $y$, the evaluation task is to assign $y$ to one of four quadrants $q \in \{Q1, Q2, Q3, Q4\}$. This formulation extends natural language inference (NLI) into a clinically grounded setting: rather than only judging whether a statement is correct, the model must also decide whether it is grounded in the patient record. By framing correctness and support jointly in a 2×2 structure, MediEval captures safety-critical confusions that cannot be identified when these dimensions are assessed separately. Although motivated by medicine, this setup provides a general template for domains where factual validity and contextual grounding are both essential. To ensure a fair comparison, all non-proprietary models are fine-tuned under an identical supervised setup with LoRA adapters and a classification head on the four quadrants, ensuring that differences in performance reflect model capabilities rather than variations in training.


\paragraph{Baseline Metrics.}  
We evaluate models on a test set $\mathcal{D}_{\text{test}} = \{(c_i, y_i, q_i)\}$ with gold quadrant labels $q_i$ and predictions $\hat{q}_i$. Overall accuracy is defined as
\begin{equation}
\text{Accuracy} = \frac{1}{|\mathcal{D}_{\text{test}}|} \sum_i \mathbb{I}(\hat{q}_i = q_i).
\end{equation}
In addition, we report macro-averaged Precision, Recall, and F1 across quadrants, together with per-quadrant F1:

\begin{equation}
F1(Q_j) = \frac{2\,\text{Prec}(Q_j)\,\text{Rec}(Q_j)}{\text{Prec}(Q_j)+\text{Rec}(Q_j)},
\quad j=1,\ldots,4.
\end{equation}

These metrics capture overall discrimination ability and expose potential biases, such as over-predicting “supported” statements.

\paragraph{Critical Error Rates.}  
Beyond global performance, clinical safety requires avoiding specific misclassifications. A model may state a correct medical fact but wrongly claim it is supported by the patient record, or elevate an incorrect statement to seemingly valid evidence. To capture these entangled risks, we introduce two error rates:

\textit{Hallucinated Support Rate (HSR).} The fraction of true but unsupported statements (Q2) misclassified as true and supported (Q1):
\begin{equation}
\text{HSR} = \frac{\sum_i \mathbb{I}(q_i = Q2 \land \hat{q}_i = Q1)}{\sum_i \mathbb{I}(q_i = Q2)}.
\end{equation}
A high HSR indicates that the model hallucinates evidence, presenting correct medical knowledge as if it were grounded in the record.  

\textit{Truth Inversion Rate (TIR).} The proportion of false but superficially supported statements (Q3) misclassified as true and supported (Q1):
\begin{equation}
\text{TIR} = \frac{\sum_i \mathbb{I}(q_i = Q3 \land \hat{q}_i = Q1)}{\sum_i \mathbb{I}(q_i = Q3)}.
\end{equation}
A high TIR indicates that the model elevates incorrect or unsafe claims to seemingly valid evidence, a critical failure mode in clinical contexts. 

\paragraph{Interpretation of Metrics.}  
These metrics disentangle complementary aspects of model behavior: \emph{Overall accuracy} and \emph{macro-F1} capture balanced classification performance. \emph{Per-quadrant F1} identifies systematic weaknesses (e.g., under-recognition of unsupported truths). \emph{HSR} and \emph{TIR} directly target the most safety-critical confusions between quadrants: hallucinating support and inverting truth.

\subsection{Counterfactual Risk-Aware Fine-Tuning}

Large language models are typically fine-tuned in two broad ways. The most direct approach is \emph{supervised fine-tuning} (SFT), where the model is trained to predict gold-standard labels with a cross-entropy loss. While simple and effective, SFT treats each label independently. In contrast, \emph{preference-based optimization} exploits comparisons between a preferred and a dispreferred output. A prominent example is \emph{Direct Preference Optimization} (DPO), which updates the model to increase the relative likelihood of preferred responses while staying close to a reference model \citep{rafailov2023direct}.

Building on these foundations and the supervised fine-tuning results established in our evaluation task, we propose \textbf{CoRFu (Counterfactual Risk-aware Fine-tuning)}, a contrastive fine-tuning strategy tailored to the clinical safety challenges revealed by MediEval. Training proceeds on preference triplets $(c, y_w, y_l)$, where $c$ is the clinical context, $y_w$ is a factually supported statement, and $y_l$ is a statement drawn from the counterfactual quadrants of MediEval. The key innovation is the \textbf{CoRFu loss function}, which augments the standard DPO objective with an asymmetric penalty that more severely punishes safety-critical errors.

Let $S$ denote the DPO preference margin:
\begin{equation}
S(c; y_w, y_l) = \beta \log [
\frac{\pi_\theta(y_w|c)\,\pi_{\text{ref}}(y_l|c)}
     {\pi_{\text{ref}}(y_w|c)\,\pi_\theta(y_l|c)}].
\label{eq:score_margin}
\end{equation}

and define the CoRFu loss as:
\begin{equation}
\mathcal{L}_{\text{CoRFu}} = -\mathbb{E} \big[ \log \sigma(S) \big] 
+ \lambda \cdot \mathbb{E}\big[\mathbb{I}(S < 0) \cdot S^2\big].
\label{eq:corfu_loss}
\end{equation}

The first term is the standard DPO objective, which encourages the model $\pi_\theta$ to prefer $y_w$ over $y_l$ relative to a reference policy $\pi_{\text{ref}}$. The second term introduces an asymmetric penalty that activates only when the model ranks $y_l$ above $y_w$ ($S<0$). Scaling this penalty quadratically in $S$ disproportionately punishes high-confidence mistakes, aligning optimization pressure with clinical safety by discouraging unsafe misinterpretations of evidence.





\section{Experimental Setup}
\label{sec:experiments}

\subsection{Dataset}
The final MediEval dataset is derived from MIMIC-IV v3.1 and consists of 2,015 unique hospital admissions, linked to 8,350 medical statements and yielding 37,144 samples (see Appendix~\ref{statistics} for distribution of relation types). The context length varies from 340 to 2,827 tokens. To prevent data leakage across splits, we retain only one admission per patient, since each patient may have multiple hospitalizations. The dataset is partitioned into training, validation, and test sets in an 80/10/10 ratio, balanced across the four quadrants. To ensure correctness, 200 test samples were randomly selected for human evaluation (details in Appendix~\ref{human_evaluation}).

\subsection{Large Language Models and Baselines}
To enable systematic comparison across model sizes and training regimes, we selected 15 LLMs spanning proprietary, open-source, and biomedical domain-specific families. 
These include \textbf{GPT} (OpenAI) \citep{openai2024gpt4technicalreport}, \textbf{LLaMA} (Meta) \citep{grattafiori2024llama}, \textbf{Mistral} (Mistral AI) \citep{jiang2024mixtralexperts}, \textbf{Qwen} (Alibaba) \citep{yang2025qwen3technicalreport}, and the biomedical models \textbf{Meditron} \citep{chen2023meditron}, \textbf{Med42} \citep{christophe2024med42}, and \textbf{ClinicalCamel} \citep{toma2023clinical}. 
For implementing CoRFu, we used the \textbf{Llama-3.1–8B-Instruct} and \textbf{Qwen3-8B} models, with direct preference optimization (DPO) serving as a baseline.

\subsection{Training Setups and Evaluation Metrics}

\paragraph{LLM evaluation with supervised fine-tuning.}
All non-proprietary models are fine-tuned under an identical supervised setup to ensure comparability. Each model is equipped with a sequence classification head and trained on the four MediEval quadrants using weighted cross-entropy loss. LoRA adapters are applied for parameter-efficient fine-tuning, with uniform hyperparameters across models (batch size, learning rate, and number of epochs). Proprietary models are evaluated in their base zero-shot generative form without fine-tuning, and their results are shown in Appendix~\ref{proprietary}.

\paragraph{Evaluation of CoRFu.}
We construct preference pairs where Q1 (supported truth) is always the preferred class. 
We explore three regimes:
\begin{itemize}
    \setlength{\itemsep}{-3pt} 
    \item \textbf{Pairwise:} contrast Q1 with a specific error type (Q2, Q3, or Q4).  
    \item \textbf{Mixed:} contrast Q1 against a pooled set of Q2/3/4 in a single stage.  
    \item \textbf{Curriculum:} train sequentially on Q1 vs.\ Q2, then Q1 vs.\ Q3, and finally Q1 vs.\ Q4, reusing the updated model at each stage.
\end{itemize}
The mixed setup exposes the model to diverse error types simultaneously, while the curriculum regime introduces them in stages of increasing difficulty, reducing interference from heterogeneous signals. By encoding MediEval’s safety perspective directly into both the loss and the training setups, CoRFu aligns model fine-tuning not only with preference correctness but also with clinical reasoning.

\paragraph{Evaluation Metrics.} Performance is measured by accuracy, macro-precision, macro-recall, macro-F1, and per-quadrant F1. Furthermore, critical error rates are quantified by the hallucinated support rate (HSR) and the truth inversion rate (TIR), for which lower values indicate better safety. 


\section{Results and Discussion}
\label{sec:results}

\begin{table*}[ht!]
\centering
\caption{MediEval benchmarking results (best results \underline{underlined}) and CoRFu results (best results in \textbf{bold}).}
\label{tab:benchmark}
\resizebox{0.95\textwidth}{!}{%
\begin{tabular}{llcccccccccc}
\toprule
& & \multicolumn{4}{c}{\textbf{Overall Performance}} & \multicolumn{4}{c}{\textbf{Per-Quadrant F1-Scores}} & \multicolumn{2}{c}{\textbf{Critical Error Rates}} \\
\cmidrule(lr){3-6} \cmidrule(lr){7-10} \cmidrule(lr){11-12}
\textbf{Type} & \textbf{Model} & \textbf{Acc.} & \textbf{Prec.} & \textbf{Rec.} & \textbf{F1} & \textbf{F1\_Q1} & \textbf{F1\_Q2} & \textbf{F1\_Q3} & \textbf{F1\_Q4} & \textbf{HSR} & \textbf{TIR} \\
\midrule
Open-source & Llama-3.1-8B-Instruct & 67.8 & 66.4 & 66.8 & 61.5 & 83.6 & 64.0 & 38.2 & 60.0 & 28.2 & 21.1 \\
  (with SFT)          & Llama-3.2-1B-Instruct & 64.1 & 57.6 & 64.3 & 50.3 & 80.9 & 0.0 & 68.1 & 52.2 & 27.3 & 15.8 \\
            & Llama-3.2-3B-Instruct & 64.1 & 61.8 & 63.8 & 61.6 & 77.1 & 53.8 & 44.8 & 70.9 & 40.9 & 31.6 \\
            & Llama-3.3-70B-Instruct & \underline{73.9} & \underline{73.3} & \underline{73.2} & \underline{70.7} & \underline{86.7} & \underline{70.0} & 65.9 & 60.0 & 21.2 & 21.1 \\
            & Mixtral-8x7B-Instruct-v0.1 & 65.4 & 69.6 & 66.0 & 63.8 & 51.4 & 68.3 & \underline{69.3} & 66.4 & \underline{20.5} & \underline{15.3} \\
            & Mistral-7B-Instruct-v0.3 & 60.5 & 61.0 & 60.7 & 59.7 & 67.8 & 55.6 & 46.2 & 69.2 & 29.1 & 26.3 \\
            & Vicuna-13B-v1.5 & 59.3 & 59.3 & 59.2 & 59.3 & 62.6 & 56.1 & 66.8 & 51.6 & 22.7 & 15.5 \\
            & Qwen3-4B & 61.7 & 61.3 & 61.7 & 61.4 & 71.7 & 49.0 & 63.3 & 61.6 & 31.8 & 26.3 \\
            & Qwen3-8B & 63.7 & 64.4 & 63.7 & 59.9 & 70.0 & 58.6 & 50.0 & 60.8 & 28.2 & 31.1 \\
            & Qwen3-32B & 62.9 & 64.6 & 63.8 & 62.1 & 72.9 & 62.6 & 51.4 & 61.3 & 28.2 & 21.1 \\
\midrule
Domain-specific & Meditron-70B & 64.4 & 62.3 & 65.2 & 68.0 & 66.2 & 64.3 & 66.7 & 74.9 & 24.8 & 26.3 \\
 (with SFT)               & Med42-70B & 65.6 & 63.3 & 65.3 & 62.8 & 58.8 & 55.8 & 58.5 & \underline{76.4} & 23.9 & 25.6 \\
                & ClinicalCamel-70B & 62.0 & 61.7 & 62.6 & 62.0 & 55.4 & 54.3 & 63.3 & 75.0 & 24.8 & 25.6 \\
\specialrule{1.2pt}{1pt}{1pt}
Llama-3.1-8B-Ins. & + DPO (Q1 vs. Q2) & 65.6 & 59.7 & 64.2 & 59.5 & 53.5 & 55.1 & 56.7 & 72.7 & 32.7 & 27.9 \\
      & + CoRFu (Q1 vs. Q2) & \textbf{76.8} & \textbf{77.2} & \textbf{77.0} & \textbf{77.9} & 73.2 & \textbf{76.6} & \textbf{78.9} & \textbf{79.9} & 18.2 & \textbf{0.0} \\
      & + CoRFu (Q1 vs. Q3) & 72.7 & 66.8 & 74.2 & 72.9 & \textbf{84.2} & 69.1 & 71.1 & 67.1 & 20.1 & 11.5 \\
      & + CoRFu (Q1 vs. Q4) & 64.5 & 74.7 & 63.6 & 69.0 & 70.6 & 67.4 & 70.0 & 68.1 & 19.9 & 17.7 \\
      & + CoRFu (Q1 vs. mix) & 69.1 & 70.2 & 69.9 & 69.6 & 67.7 & 68.6 & 72.1 & 71.7 & 23.1 & 12.9 \\
      & + CoRFu (Curriculum) & 64.5 & 75.0 & 63.6 & 70.6 & 70.6 & 70.5 & 72.3 & 68.8 & 19.1 & 3.5 \\
\midrule
Qwen3-8B & + DPO (Q1 vs. Q2) & 51.0 & 51.6 & 50.9 & 51.1 & 51.2 & 49.1 & 56.7 & 47.4 & 22.7 & 4.4 \\
      & + CoRFu (Q1 vs. Q2) & 70.7 & 71.5 & 71.1 & 71.0 & 65.1 & 66.7 & 78.0 & 74.3 & 21.8 & \textbf{0.0} \\
      & + CoRFu (Q1 vs. Q3) & 61.1 & 52.0 & 59.9 & 53.0 & 52.9 & 55.3 & 49.3 & 54.4 & 22.3 & 8.3 \\
      & + CoRFu (Q1 vs. Q4) & 60.3 & 55.2 & 59.6 & 50.7 & 54.3 & 50.1 & 51.3 & 46.7 & 18.9 & 7.6 \\
      & + CoRFu (Q1 vs. mix) & 59.3 & 60.7 & 60.3 & 59.3 & 53.3 & 52.4 & 68.9 & 62.7 & 19.3 & 13.3 \\
      & + CoRFu (Curriculum) & 48.2 & 46.0 & 48.1 & 47.7 & 40.8 & 45.3 & 41.1 & 49.2 & \textbf{17.3} & 14.5 \\
\bottomrule
\end{tabular}%
}
\end{table*}

\subsection{MediEval Benchmarking Results}
\paragraph{Overall Performance.}
Table~\ref{tab:benchmark} reports macro metrics (accuracy, precision, recall, F1) and safety scores (HSR, TIR) on MediEval. Overall performance is limited given the clinical reasoning requirements, with accuracies ranging from 59.3\% to 73.9\% and macro F1 from 50.3\% to 70.7\% across base models. The strongest model is \texttt{Llama-3.3-70B-Instruct}, which achieves the highest overall results in accuracy (73.9\%), precision (73.3\%), recall (73.2\%), and F1 (70.7\%). Among smaller models, \texttt{Llama-3.1-8B-Instruct} is competitive, reaching 61.5\% macro F1. Although pretrained on clinical corpora, biomedical LLMs do not dominate. These results suggest that lexical familiarity and domain exposure alone are insufficient; fine-grained reasoning to disentangle factual correctness from contextual grounding requires explicit supervision.

\paragraph{Per-Quadrant Performance.}
Performance is uneven across the four quadrants, but consistent with their expected difficulty. 
\textbf{Q1} (Supported Fact) is easiest (best 86.7 by \texttt{Llama-3.3-70B}), consistent with models’ strength on recognized truthful patterns in-context. \textbf{Q2} (Plausible Contradiction) is harder, with the best model below 70\%, showing that LLMs often treat true but irrelevant statements as supported. \textbf{Q3} (Relational Mismatch) stresses relation-level reasoning; the best score reaches only 69.3, showing limited resistance to superficially coherent recombinations. Domain-specific models excel on \textbf{Q4} (Erroneous Information) (74.9 to 76.4\%), likely because clinical pretraining gives them stronger priors for rejecting clearly false medical statements, though their broader reasoning across quadrants remains limited. Notably, even within a single model (e.g., \texttt{Llama-3.1-8B-Instruct}), performance varies sharply across quadrants, showing that strength in one type of reasoning does not transfer to others. This unevenness highlights persistent weaknesses in how models combine factuality with patient grounding.

\paragraph{Safety-Critical Errors.}
Safety-critical metrics expose risks not captured by aggregate F1. \texttt{Mixtral-8x7B} consistently achieves the lowest error rates, with HSR (Q2$\!\to$Q1) of 20.5\% and TIR (Q3$\!\to$Q1) of 15.3\%, whereas \texttt{Llama-3.2-3B} exhibits the highest risk on both metrics (HSR 40.9\%, TIR 31.6\%). This indicates that Mixtral is the safest model with respect to HSR/TIR, even though it does not attain the highest macro F1. In contrast, \texttt{Llama-3.3-70B} achieves the best macro F1 (70.7\%) yet fails to minimize HSR/TIR (21.2/21.1), demonstrating that higher accuracy does not necessarily translate into safer clinical reasoning. Overall, the divergence between F1 and safety metrics shows that models can perform well on average while still making risky errors.

\subsection{CoRFu Evaluation Results}
\paragraph{Overall Effects.}
Across both backbones, CoRFu outperforms the base SFT and DPO models, but the strongest gains concentrate in a single pairing. On \texttt{Llama-3.1-8B-Instruct}, \textbf{Q1 vs.\ Q2} achieves the best results at 76.8\% accuracy and 77.9\% macro F1 (+16.4 F1 over SFT), while other variants are lower (e.g., Q1 vs.\ Q3 at 72.9\% F1, Q1 vs.\ mix at 69.6\%, curriculum at 70.6\%). For \texttt{Qwen3-8B}, the pattern is sharper: only \textbf{Q1 vs.\ Q2} yields a substantial aggregate improvement (70.7\% Acc, 71.0\% F1; +11.1 F1 over base), whereas Q1 vs.\ Q3, Q1 vs.\ Q4, and curriculum underperform on macro metrics. This shows that contrasting supported with unsupported truths provides the most reliable signal for improving overall reasoning, while other pairings mainly target specific weaknesses.

\paragraph{Per-Quadrant Improvements.}
CoRFu lifts per-quadrant F1 scores, but the biggest gains come from targeted pairings rather than curriculum. With the \texttt{Llama-3.1-8B-Instruct} backbone, \textbf{Q1 vs.\ Q2} achieves the best balance across quadrants (Q1/Q2/Q3/Q4 = 73.2/76.6/78.9/79.9), improving both factual grounding (Q2) and overall accuracy.  
Targeted objectives produce predictable effects: training on (Q1 vs.\ Q2) substantially boosts Q2 F1 (+12.6 over base), while (Q1 vs.\ Q3) increases Q3 F1. These results show that targeted pairings not only correct specific weaknesses but also provide complementary improvements across quadrants.

\paragraph{Safety-Critical Reductions.}
Improvements in accuracy do not automatically translate into safety, but CoRFu reduces both HSR and TIR compared to SFT and DPO.  
On \texttt{Llama-3.1-8B}, the (Q1 vs.\ Q2) run achieves the lowest HSR at 18.2\% and eliminates TIR entirely.  
On \texttt{Qwen-8B}, curriculum yields the lowest HSR at 17.3\%, while (Q1 vs.\ Q2) again minimizes TIR to 0.0. These results show that different contrastive objectives selectively mitigate different risks, confirming that aggregate accuracy and safety-critical errors must be disentangled and addressed with tailored supervision.

\paragraph{Ablation Study.}
We conducted an ablation study for the value of $\lambda$. When $\lambda=0$, the method reduces to vanilla DPO, with performance reported in Table~\ref{tab:benchmark}. Values between 0.5 and 1.0 yield the best macro-F1 while significantly lowering HSR and TIR. For $\lambda > 1.0$, over-penalization occurs, leading to a decrease in macro-F1 and an increase in error rates. See Appendix~\ref{further} \& \ref{qualitative}
 for more analyses.

\paragraph{Overall Insight.}
Taken together, the results show that contrastive objectives shape model behavior in complementary ways. Q1–Q2 training provides the strongest overall gains, improving accuracy, macro F1, and grounding while driving HSR to its lowest levels. Q1–Q3 supervision instead strengthens supported reasoning and sharply lowers TIR, though with smaller aggregate benefits. Curriculum is less competitive on headline metrics but can reduce HSR further for some backbones. Overall, these results demonstrate that contrastive fine-tuning can be directed to improve specific weaknesses, and that MediEval makes such effects visible by disentangling accuracy from safety-critical errors.

\section{Conclusion}

We introduced \textbf{MediEval}, a benchmark that jointly evaluates factual verification and patient-contextual reasoning by linking biomedical ontologies with real-world patient records. Our results show that current LLMs struggle with quadrant-level reasoning and make safety-critical errors not reflected in aggregate accuracy. To mitigate these risks, we proposed \textbf{Counterfactual Risk-Aware Fine-tuning (CoRFu)}, which reduces unsafe misclassifications through asymmetric preference optimization.

For future work, MediEval could be extended to other clinical tasks such as question answering, adapted to safety-critical domains beyond medicine, and integrated with methods such as retrieval-augmented generation or models designed for temporal patient reasoning. We hope this work provides a foundation for more rigorous and risk-aware evaluation of LLMs.


\section*{Limitations}
\textbf{Dataset size and coverage.} MediEval is constructed from approximately 2k admissions in MIMIC-IV, yielding around 37k statements. This scale reflects a deliberate design choice to prioritize clinical plausibility, careful quality control, and strict prevention of patient-level information leakage, which are particularly important in safety-critical medical evaluation. The dataset construction pipeline is fully automated and released with the codebase, making MediEval readily extensible to larger cohorts, additional institutions, or alternative clinical subsets as future work.

\textbf{Annotation and validation.} A subset of the test set was independently reviewed by medically trained annotators, with GPT-5 used as an auxiliary reference to inspect disagreements. High agreement suggests that the ontology-grounded construction leads to largely unambiguous labels. While broader expert validation could further strengthen confidence, the reliance on structured biomedical knowledge and deterministic generation reduces subjectivity compared to fully manual annotation.

\textbf{Data and ontology noise.} As with any benchmark grounded in real-world clinical data and large biomedical ontologies, MediEval inherits noise and inconsistencies from MIMIC-IV and resources such as UMLS, SNOMED CT, and RxNorm. We mitigate these effects through ontology-guided plausibility checks, constrained multi-hop traversal, and explicit grounding in patient context. Further advances in ontology curation and clinical data quality are likely to directly benefit benchmarks such as MediEval.


\section*{Acknowledgments}
The authors acknowledge the financial support by the Federal Ministry of Research, Technology and Space of Germany and by Sächsische Staatsministerium für Wissenschaft, Kultur und Tourismus in the programme Center of Excellence for AI-research „Center for Scalable Data Analytics and Artificial Intelligence Dresden/Leipzig“, project identification number: ScaDS.AI.

The authors acknowledge the computing time made available to them on the high-performance computer at the NHR Center of TU Dresden. This center is jointly supported by the Federal Ministry of Research, Technology and Space of Germany and the state governments participating in the NHR (www.nhr-verein.de/unsere-partner).

\bibliography{custom}

\newpage
\appendix
\section*{Appendices}
\section{Real Samples for Each Quadrant}
\label{sec:appendixa}

This appendix provides representative examples of MediEval samples from each of the four quadrants (see Figure~\ref{fig:example} for Q2). Each example consists of a patient context extracted from a real MIMIC-IV discharge summary, a constructed medical statement, and its corresponding quadrant label.

\begin{figure}[!ht]
    \centering
    \includegraphics[width=1\linewidth]{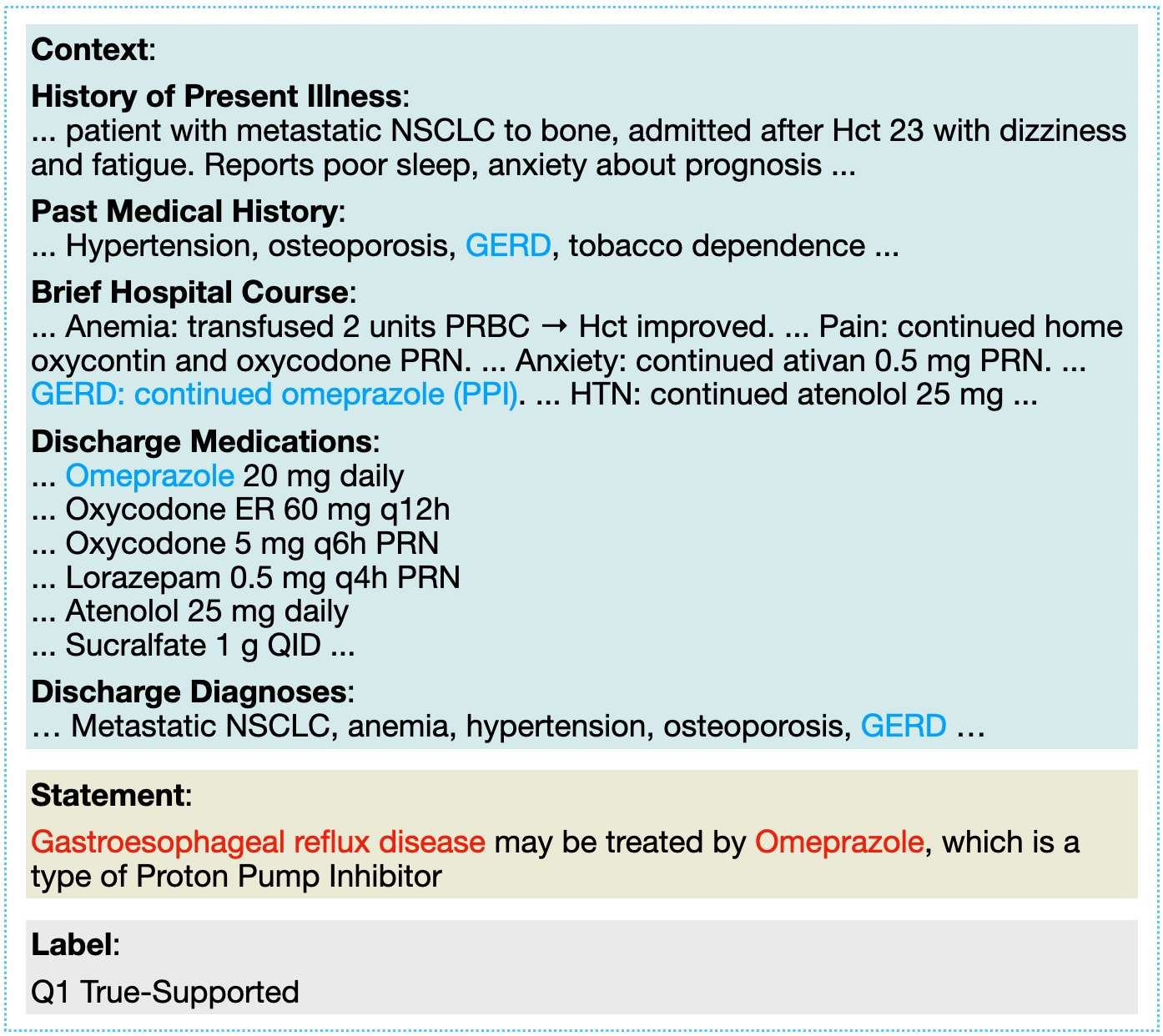}
    \caption{Example of statement verification against patient records (Quadrant 1: True-Supported). The statement is medically correct, and GERD is indeed treated with omeprazole, making it true and supported.}
    \label{fig:exampleq1}
\end{figure}

\begin{figure}[!ht]
    \centering
    \includegraphics[width=1\linewidth]{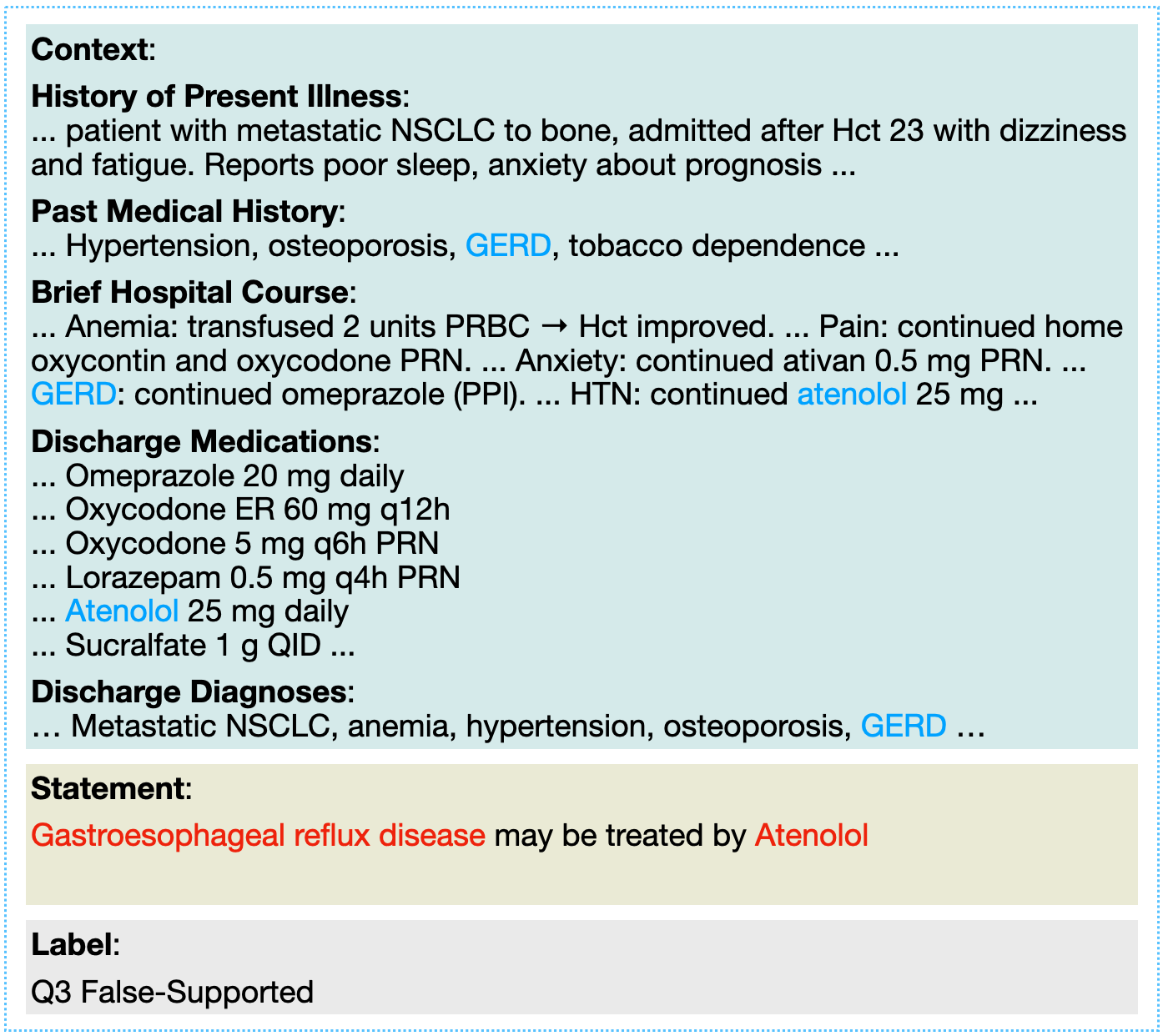}
    \caption{Example of statement verification against patient records (Quadrant 3: False-Supported). The crafted statement is medically incorrect, as Atenolol is indicated for hypertension. Since both GERD and Atenolol appear in the patient context, the statement may seem supported, even though it is false.}
    \label{fig:exampleq3}
\end{figure}

\begin{figure}[!t]
    \centering
    \includegraphics[width=1\linewidth]{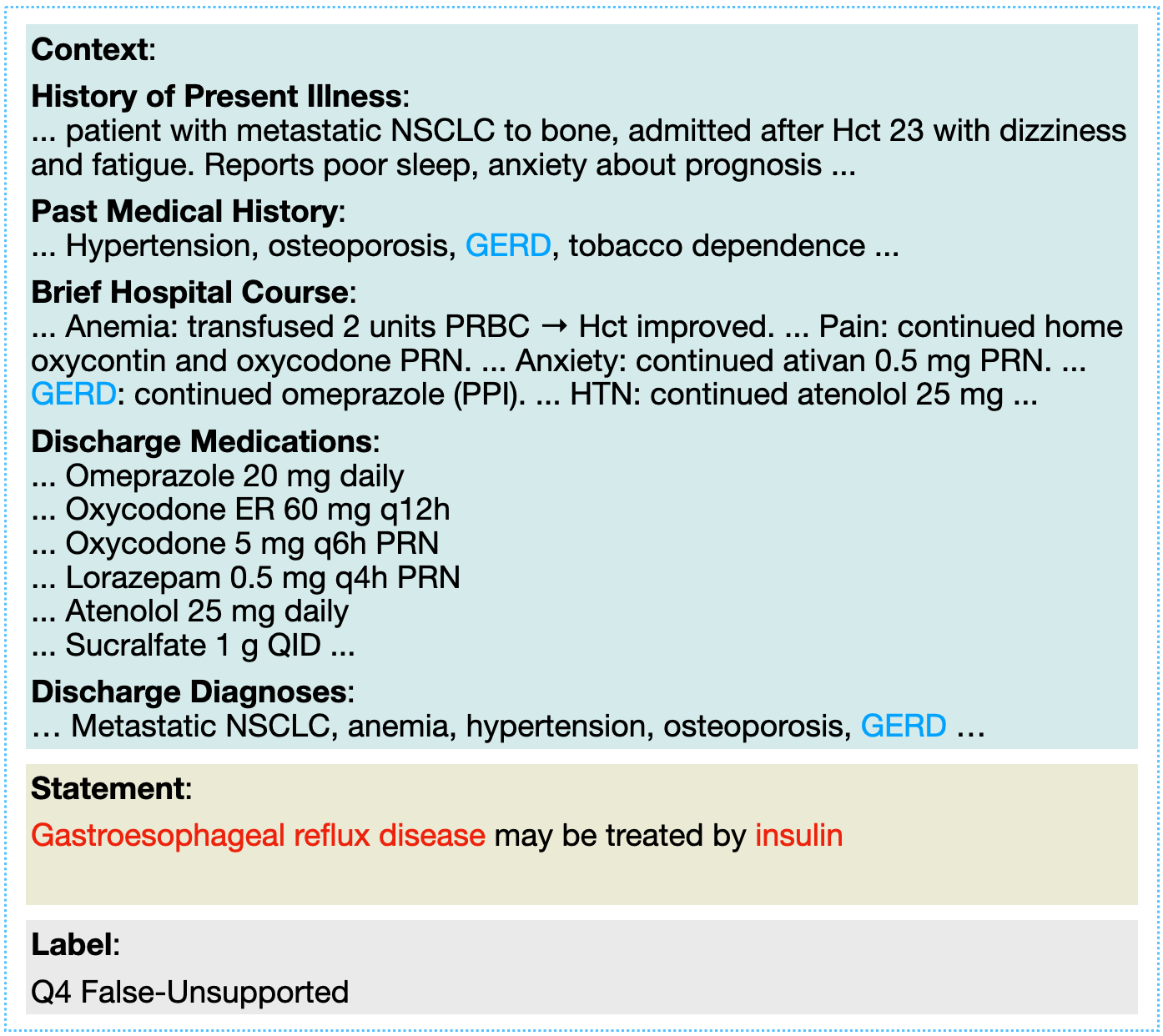}
    \caption{Example of statement verification against patient records (Quadrant 4; False-Unsupported). The statement is medically incorrect, since insulin is not indicated for GERD. Moreover, insulin does not appear in the patient context, making the statement false and unsupported.}
    \label{fig:exampleq4}
\end{figure}

\FloatBarrier
\section{Distribution of Relation Types}
\label{statistics}
This appendix reports the distribution of biomedical relation types used in MediEval. Relations are derived from UMLS and associated vocabularies after semantic normalization and ontology-guided traversal. The distribution reflects the clinical diversity of the benchmark, covering treatment, diagnostic, pharmacologic, and associative relations.

\begin{table}[!hb]
\centering
\caption{Distribution of relation types in the dataset}
\label{tab:relation_statistics}
\resizebox{0.65\linewidth}{!}{%
\begin{tabular}{l r}
\hline
\textbf{Relation Type} & \textbf{Count} \\
\hline
treats & 17{,}540 \\
may\_be\_treated\_by & 4{,}552 \\
related\_to & 4{,}008 \\
classified\_as & 3{,}774 \\
contraindicated\_class\_of & 1{,}250 \\
subset\_includes & 1{,}568 \\
has\_pharmacologic\_class & 854 \\
diagnoses & 738 \\
prevents & 748 \\
causes & 808 \\
associated\_with & 588 \\
authorized\_value & 282 \\
co-occurs\_with & 130 \\
defining\_characteristic\_of & 230 \\
manifestation\_of & 56 \\
sign\_or\_symptom\_of & 12 \\
may\_be\_diagnosed\_by & 4 \\
part\_of & 2 \\
\hline
\textbf{Total} & \textbf{37{,}144} \\
\hline
\end{tabular}%
}
\end{table}

\begin{table*}[!htb]
\centering
\caption{MediEval benchmarking results of Proprietary GPT Models.}
\label{tab:proprietary}
\resizebox{0.85\textwidth}{!}{%
\begin{tabular}{lcccccccccc}
\toprule
& \multicolumn{4}{c}{\textbf{Overall Performance}} & \multicolumn{4}{c}{\textbf{Per-Quadrant F1-Scores}} & \multicolumn{2}{c}{\textbf{Critical Error Rates}} \\
\cmidrule(lr){2-5} \cmidrule(lr){6-9} \cmidrule(lr){10-11}
\textbf{Model} & \textbf{Acc.} & \textbf{Prec.} & \textbf{Rec.} & \textbf{F1} & \textbf{F1\_Q1} & \textbf{F1\_Q2} & \textbf{F1\_Q3} & \textbf{F1\_Q4} & \textbf{HSR} & \textbf{TIR} \\
\midrule
GPT-5 & 45.7 & 36.4 & 46.7 & 30.4 & 68.3 & 53.3 & 0.0 & 0.0 & 19.3 & 0.0 \\
GPT-4o & 49.3 & 31.6 & 50.0 & 29.1 & 66.3 & 50.0 & 0.0 & 0.0 & 46.7 & 0.0 \\
\bottomrule
\end{tabular}%
}
\end{table*}

\section{Human Evaluation}
\label{human_evaluation}
We conducted human validation on 200 balanced samples. Two medically trained annotators labeled each item independently. In cases of disagreement, GPT-5 was consulted as an auxiliary signal for sanity checking, but final labels were determined by the human annotators. Agreement was 97\% (194/200), with Cohen’s $ kappa \approx 0.96$, indicating strong agreement. All disagreements occurred between Q2 and Q3, which is one of the subtle distinctions that the benchmark is designed to probe. Because our extraction pipeline is fully ontology-driven and deterministic, the errors stem from noise or inconsistencies in the underlying ontologies or EHR data.

\section{Implementation Details}

All training was performed on an NVIDIA L40S GPU with 46 GB memory. All reported results are averaged over 3 random seeds (42, 43, 44). For the MediEval supervised baselines, we fine-tuned models for 3 epochs with per-device batch size 4, gradient accumulation 8, maximum sequence length 4096, learning rate $2 \times 10^{-5}$, weight decay 0.01, and warmup ratio 0.03. When enabled, LoRA used rank $r = 8$, $\alpha = 16$, and dropout $0.05$. For the CoRFu experiments, we fine-tuned with maximum sequence length 4096, per-device batch size 1, gradient accumulation 16 (effective batch size 16), and trained for 1 epoch. We used the AdamW optimizer with learning rate $1 \times 10^{-4}$, $\beta_1 = 0.9$, $\beta_2 = 0.999$, weight decay 0.01, warmup ratio 0.05, and gradient clipping at 1.0. LoRA was configured with rank $r = 16$, $\alpha = 32$, and dropout $0.1$. The CoRFu loss used $\beta = 0.1$ and $\lambda = 0.5$.

\section{Evaluation of Proprietary GPT Models}
\label{proprietary}
GPT-4o \cite{hurst2024gpt} and GPT-5 \cite{singh2025openai} are evaluated under a realistic clinical setting, following their typical usage: zero-shot inference without task-specific fine-tuning. This reflects practical deployment constraints, where adapting proprietary models is often infeasible. Our goal is not a direct head-to-head comparison, but rather to highlight two complementary observations: (1) the task remains challenging even for frontier models without adaptation, and (2) smaller models equipped with targeted supervision (SFT/CoRFu) can outperform general-purpose LLMs on safety-critical dimensions.

For proprietary models, the prompt consists of the following components: (1) context, (2) statement, (3) definitions of the four quadrants, (4) an instruction to select exactly one label, and (5) the list of candidate labels.

Overall, proprietary models struggle on this task. GPT models achieve only 29.1\% (GPT-4o) and 30.4\% (GPT-5) macro-F1. Since both models are evaluated in their base generative form without task-specific adaptation, this performance reflects both the difficulty of MediEval and the lack of alignment to the evaluation protocol.

In terms of safety errors, HSR (Q2$\!\rightarrow$Q1) ranges from 19.3\% for GPT-5 to 46.7\% for GPT-4o. Notably, proprietary models never predict Q3, resulting in zero TIR. This behavior indicates miscalibration rather than genuine safety awareness.

\section{Ablation Study}
\label{further}

\begin{figure}[!tbh]
  \centering
  \includegraphics[width=0.83\linewidth]{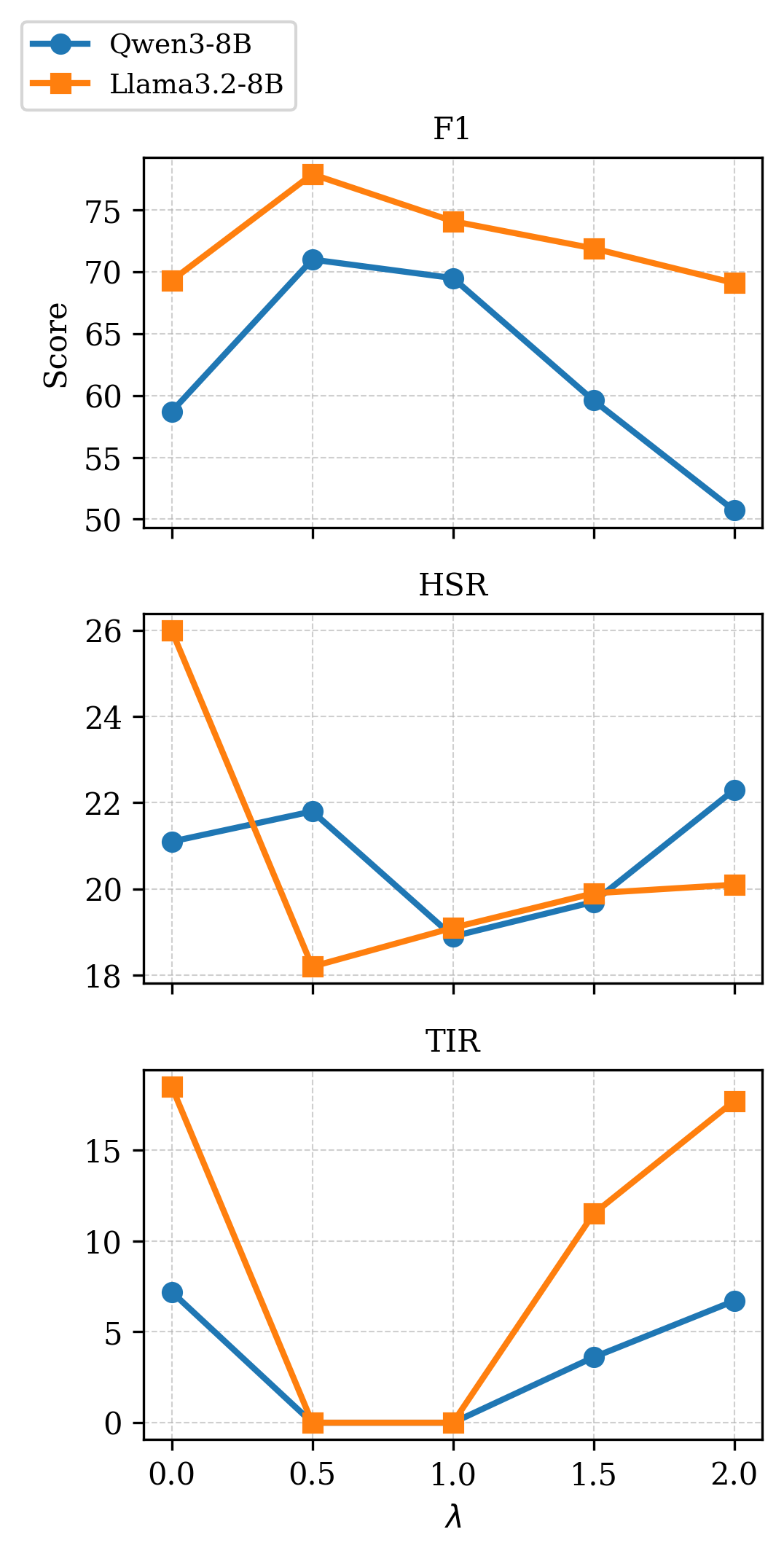}
  \caption{Effect of the regularization coefficient $\lambda$ on macro-F1, HSR, and TIR for Qwen3-8B and Llama-3.1-8B-Instruct under Q1 vs. Q2 CoRFu training.}
  \label{fig:lambda_metrics}
\end{figure}

Figure~\ref{fig:lambda_metrics} illustrates the effect of the regularization coefficient $\lambda$ on performance and error rates, under Q1 vs. Q2 CoRFu training. For both models, moderate values of $\lambda$ (around $0.5$--$1.0$) achieve the best trade-off: macro-F1 is maximized, while HSR and TIR are substantially reduced compared to $\lambda=0$, which corresponds to vanilla DPO. As $\lambda$ increases further, performance degrades and error rates rise, indicating over-penalization of negative preference margins. This ablation is necessary because $\lambda$ directly controls the strength of the corrective penalty introduced by CoRFu, balancing preference optimization against error suppression. In contrast, we fix the DPO temperature $\beta$, since it only rescales the preference margin and its effect can be absorbed into $\lambda$.

\section{Qualitative Evaluation}
\label{qualitative}
We further analyze whether model behavior correlates with input characteristics such as context length or relation type, across all evaluated settings, including SFT and CoRFu. We observe no consistent correlation between any of the reported evaluation metrics and context length, nor systematic performance differences across relation types for any model. These results indicate that the observed performance trends are not driven by input complexity or relation semantics, but instead reflect differences in training objectives and supervision strategies.

Instead, errors predominantly arise from confusions between Q2 and Q1, indicating difficulties in distinguishing between partially incorrect and fully incorrect cases. We provide a qualitative analysis of these failure modes below.

\begin{tcolorbox}[
  title=Qualitative Example - hadm\_id: 24480054,
  breakable,
  colback=gray!5,
  colframe=black,
  boxrule=0.4pt,
  left=1.5mm,
  right=1.5mm,
  top=1mm,
  bottom=1mm
]
\setlength{\parskip}{0.1em}

\textbf{Statement.} \emph{Gastroesophageal reflux disease may be treated by aluminum hydroxide, which is a type of antacid.}

\medskip
\textbf{Context.}
\begin{lstlisting}
History of Present Illness:
___ is a G4P3 female with a history of thrombophilia and cerebrovascular accident who presented to an outside hospital with worsening epigastric pain without overt signs of labor. She was admitted in early labor and subsequently delivered her baby on the morning of ___. However, her abdominal pain continued to worsen during hospitalization.

She has a history of gastroesophageal reflux disease previously treated effectively with pantoprazole, later transitioned to omeprazole due to insurance changes, with recurrence of symptoms. During the first trimester of pregnancy, she experienced worsening epigastric pain with significant nausea and vomiting. After the first trimester, nausea and vomiting improved, and her heartburn and epigastric pain also improved. In the week prior to delivery, her symptoms worsened again, with pain primarily localized to the epigastrium and exacerbated by movement. She attempted over-the-counter Tums and Pepcid without relief.

At the outside hospital, she underwent abdominal ultrasound and CT imaging demonstrating pancreatitis. She was noted to have leukocytosis and received imipenem prior to transfer. She received at least 4L of normal saline and voided at least 350cc. She reported bilious emesis but denied ongoing nausea.

Her delivery was uncomplicated. Her newborn son, ___, is healthy. No estimated blood loss was recorded, and delivery was vaginal. She continued to experience suprapubic pain.

On arrival to the MICU, vital signs were: temperature 97.9F, blood pressure 156/97, heart rate 107, respiratory rate 16, and SpO2 97% on nasal cannula. She reported abdominal pain that was tolerable and improved with hydromorphone. She denied shortness of breath. She had not passed gas, and her last bowel movement was ___.

Past Medical History:
- Stroke associated with Depo-Provera, without residual deficits
- Thrombophilia (MTHFR deficiency)
- Hepatitis C (cleared virus; viral load undetectable, antibody positive; no IVDU; history of blood transfusion ___)
- Ruptured appendix ___
- Morbid obesity
- Gastroesophageal reflux disease
- Pregnancy in ___ with possible fetal alcohol syndrome, ADHD, and bipolar disorder; no issues in ___ or ___ pregnancies

Brief Hospital Course:
Severe Acute Pancreatitis: The patient was admitted to the MICU and kept NPO. MRCP demonstrated greater than 30% pancreatic necrosis with a hemorrhagic component. A nasojejunal tube was placed, and tube feeds were initiated. Initial intolerance improved with cycling. No drainable fluid collections were identified. Gastroenterology followed throughout admission. She was discharged on a clear liquid diet with plans to advance to a low-fat diet. Persistent left upper quadrant pain gradually improved.

Thrombophilia: The patient has a history of MTHFR deficiency and prior stroke. A new non-occlusive portal vein thrombosis was identified during admission. Therapeutic enoxaparin was initiated after stabilization of hemorrhagic pancreatitis, with a planned duration of ___ months and hematology follow-up.

Postpartum Status: The patient had an uncomplicated vaginal delivery of a healthy infant on ___. Social work was involved due to limited home support.

Acute Kidney Injury: An acute kidney injury was present on admission, likely secondary to pancreatitis, and resolved with intravenous fluid resuscitation.

Ileus and Diarrhea: An initial ileus was treated conservatively. The patient later developed tube-feed--associated diarrhea. Clostridioides difficile testing was negative. Symptoms improved with as-needed loperamide.

New-Onset Diabetes Mellitus: Hyperglycemia developed after initiation of tube feeds, likely secondary to severe pancreatitis. The patient was started on an insulin sliding scale. Due to clinical complexity, she was discharged on a Humalog sliding scale only and requires outpatient reassessment.

Inactive Issues:
Gastroesophageal reflux disease: continue pantoprazole.

Discharge Diagnoses:
- Acute severe pancreatitis
- Hyperglycemia, likely new-onset diabetes mellitus
- Portal vein thrombosis
- Acute kidney injury, resolved
\end{lstlisting}

\medskip
\textbf{Label.} True-Supported 
\end{tcolorbox}

The statement is medically correct in general; however, the specific treatment (aluminum hydroxide) is not mentioned in the clinical context. While GERD is documented and treated with pantoprazole and other antacids, the generated statement introduces an unsupported medication, and is therefore labeled as Q2.

Almost all models fail on this example due to a combination of lexical bias and treatment-level over-generalization. The presence of strong surface cues such as GERD, reflux, and references to antacid use biases models toward conceptual matching, leading them to infer support even when the specific medication (aluminum hydroxide) is not mentioned. 

In addition, models tend to collapse distinctions within therapeutic classes, treating different antacids as interchangeable. This error is further exacerbated by clinical complexity rather than context length: the note is dominated by severe acute pancreatitis and post-partum complications, while GERD appears as a secondary issue, making fine-grained verification of drug-level evidence particularly challenging. Together, these factors reveal a persistent difficulty in jointly verifying medical correctness and strict contextual grounding.

\end{document}